\documentclass[conference]{IEEEtran}
\IEEEoverridecommandlockouts
\usepackage{amsmath}
\usepackage{algorithm}
\usepackage{algpseudocode}
\usepackage{amsthm}
\usepackage{graphicx}
\usepackage{xcolor}

\usepackage[utf8]{inputenc} 
\usepackage[T1]{fontenc}    
\usepackage{hyperref}       
\usepackage{multirow}       
\hypersetup{
    colorlinks=true,
    linkcolor=blue,
    citecolor=blue,
    filecolor=magenta,
    urlcolor=blue,
}
\usepackage{url}            
\usepackage{booktabs}       
\usepackage{amsfonts}       
\usepackage{nicefrac}       
\usepackage{tabularx}
\usepackage{booktabs}
\usepackage[noadjust]{cite}
\usepackage{enumitem}
\usepackage{array}
\usepackage{ulem}
\usepackage{soul}
\usepackage{amsmath}
\usepackage{fancyhdr}
\usepackage{float}  
\usepackage{listings}
\usepackage{soul}
\usepackage{scrextend}

\usepackage{newfloat}
\DeclareFloatingEnvironment[%
   fileext=loe,
   name=Exhibit
]{exhibit}

\makeatletter
\DeclareRobustCommand*\cal{\@fontswitch\relax\mathcal}
\makeatother

\def\BibTeX{{\rm B\kern-.05em{\sc i\kern-.025em b}\kern-.08em
    T\kern-.1667em\lower.7ex\hbox{E}\kern-.125emX}}

\makeatletter
\newcommand{\linebreakand}{%
  \end{@IEEEauthorhalign}
  \hfill\mbox{}\par
  \mbox{}\hfill\begin{@IEEEauthorhalign}
}
\makeatother

\begin{document}

\title{Converting Expert Deliberation into Financial Signals Through A Context-Aware NLP Pipeline}

\author{
\IEEEauthorblockN{[Author names omitted for review]}
\IEEEauthorblockA{[Affiliations omitted for review]}
}

\author{

\IEEEauthorblockN{Vivek Batra}
\IEEEauthorblockA{
\textit{Franklin Templeton Investments}\\
Mumbai, India \\
{\small vivek.batra@franklintempleton.com}}
\and
\IEEEauthorblockN{Kristin Chen}
\IEEEauthorblockA{
\textit{Blend360}\\
Columbia, USA \\
{\small kristin.chen@blend360.com}}
\and
\IEEEauthorblockN{Sanjiv R. Das}
\IEEEauthorblockA{
\textit{Santa Clara University}\\
Santa Clara, USA \\
{\small srdas@scu.edu}}
\linebreakand
\IEEEauthorblockN{Samuel Judge}
\IEEEauthorblockA{
\textit{Franklin Templeton Investments}\\
Boston, USA \\
{\small samuel.judge@franklintempleton.com}}
\and
\IEEEauthorblockN{Harshad Khadilkar}
\IEEEauthorblockA{
\textit{Franklin Templeton Investments}\\
Mumbai, India \\
{\small harshad.khadilkar@franklintempleton.com}}
\and
\IEEEauthorblockN{Sukrit Mittal}
\IEEEauthorblockA{
\textit{Franklin Templeton Investments}\\
Hyderabad, India \\
{\small sukrit.mittal@franklintempleton.com}}
\linebreakand
\IEEEauthorblockN{Amir Nasrollahzadeh}
\IEEEauthorblockA{
\textit{Blend360}\\
Boston, USA \\
{\small amir.nasrollahzadeh@blend360.com}}
\and
\IEEEauthorblockN{Daniel Ostrov}
\IEEEauthorblockA{
\textit{Santa Clara University}\\
Santa Clara, USA \\
{\small dostrov@scu.edu}}
\and
\IEEEauthorblockN{Jacob Sisk}
\IEEEauthorblockA{
\textit{Franklin Templeton Investments}\\
San Ramon, USA \\
{\small jacob.sisk@franklintempleton.com}}

}

\maketitle

\begin{abstract}
We introduce the CDSP (context-conditional deliberation signal pipeline), converting an investment committee's meeting transcripts into structured predictive features. CDSP segments the meeting transcripts into topical chunks, assigns asset-class context labels using a large language model (LLM), maps financial keywords to a pre-determined taxonomy of labels, and constructs complementary features: sentiment polarity and mention frequency. This feature engineering framework is applied to a dataset spanning 48 monthly committee meetings to predict if global equities will perform better or worse than global bonds in the following month. In experiments with engineered features, raw transcript text, sentence embeddings, and combined representations, the prediction accuracy ranges from 62\% to 73\%, compared to always choosing stocks, which outperforms bonds 60.4\% of the time. The best (73\% accurate) model combines sentence embeddings with engineered CDSP features, achieving a 0.73 F1 score (although this is not statistically significant compared to always choosing stocks). Sentiment carries a stronger signal than mention frequency for several taxonomy categories. These findings suggest that experts' deliberations may contain forward-looking information that context-aware NLP can extract.

\end{abstract}


\section{Introduction}

Firms rarely make high-stakes decisions by simply observing numbers and acting on them. The underlying subjectivity, including the debate over upside versus downside and the comparison of alternative scenarios, helps them form a collective view on the best path forward. Investment committees or portfolio research groups operate in a similar manner on a regular basis. They interpret uncertain information, debate future possibilities, and decide how to position themselves before that future is known. These discussions are often recorded for review, compliance and institutional memory, effectively creating an archive of expert reasoning. With the availability of Large Language Model (LLM)-based tools, there is now an opportunity to systematically extract the thought process behind those decisions: for example, how a geopolitical condition was weighed against an offsetting macroeconomic condition, and how well the resulting decision ultimately performed. 

At first glance, it may seem straightforward to feed such historical archives into LLMs with large context windows, and query them on an as-needed basis. However, this raises questions that are particularly relevant in context-sensitive investment applications:
\begin{itemize}
    \item Do investment committee transcripts contain structured predictive signals about market performance in the short term? 
    \item Does the meaning and the prediction power of financial language change depending on where and in which context it appears within the discussion? 
\end{itemize}
Although the LLM may provide answers to both questions, uncertainty remains because LLMs are largely black boxes. This paper attempts to address that uncertainty by treating the historical archive as a structured data source and evaluating its empirical merit in making predictions. We study whether the language used in investment committee meetings contains structured signals about the direction of equities versus bonds in the following month. We assess whether modern natural language processing (NLP) can transform expert discussions into interpretable features for supervised learning models.

In the context of this paper, an investment committee meets each month to discuss macroeconomic conditions, outlooks for asset classes like equities and bonds, and portfolio positioning across these assets. The committee's objective is to form a forward-looking view about relative opportunities and risks. For example, a committee may discuss whether equity valuations are attractive relative to bond yields, or whether macroeconomic risks justify reducing exposure to riskier assets. These conversations therefore contain opinions that are guided by perceived financial patterns. 

Our focus is the simple prediction problem of whether or not equities will outperform bonds in the following month. This captures allocation decision information in a simple form. 
The key questions here are whether the committee's discussion contains enough information to help predict this, and, more importantly, whether NLP techniques can use this information to generate meaningful predictions. 

To answer these key questions, we introduce the Context-conditional Deliberation Signal Pipeline (CDSP). CDSP cleans and segments meeting transcripts into topic-based chunks. It then uses an LLM to assign to each chunk one or more context labels from the set $\mathcal{C}=$\{equity, bonds, cross-asset, macroeconomic-outlook, meeting-logistic, other\}. From a vocabulary of 2416-keyword domains, financial keywords are extracted from each chunk and mapped to a curated taxonomy of 15 subcategories contained within four investment attribute categories (given in Table~\ref{tab:taxonomy}). For each topic and context label, we construct two feature categories: \textit{sentiment}, which measures how positive or negative the surrounding language is, and \textit{mention frequency or density}, which measures how much attention the committee gives to that topic. The key idea is \textit{context-conditioning}. The same financial phrase may have a different meaning depending on where it appears in the meeting. For example, discussion of valuation or credit conditions during an equity-focused segment can carry a different signal than the same language during a macroeconomic overview. 


In this paper, we compare three ways of representing the transcripts: (1) using engineered semantic features produced by CDSP, (2) using the transcript text directly, and (3) using sentence embeddings, which provide dense numerical representations. We also test whether combining engineered features with embeddings improves performance. Across these approaches, the classification accuracy for the next-month market movement ranges from 62\%--73\%, where the best results come from combining sentence embeddings with engineered CDSP features, achieving 73\% accuracy and a 0.73 F1 score under repeated stratified cross-validation. 

The remainder of the paper is structured as follows. Section~\ref{sec:background} reviews related work. Section~\ref{sec:data} describes the dataset. Section~\ref{sec:pipeline} details the CDSP framework. Section~\ref{sec:features} presents feature analysis. Section~\ref{sec:model} describes the predictive models and results. Section~\ref{sec:discussion} discusses implications and limitations. Section~\ref{sec:conclusion} concludes.

\section{Related Work}
\label{sec:background}

\subsection{Text-Based Prediction in Finance}
NLP has been actively explored for financial applications such as sentiment analysis, financial forecasting, portfolio management, risk management, financial narrative processing, and explainable AI \cite{du2025natural}. Most studies in this area explored using public or corporate text, including news, filings, annual reports, earnings calls, analyst reports, social media, and online forums (like X, Reddit, etc.). These sources have been shown to contain information relevant to returns, volatility, and investor reactions \cite{kogan2009predicting, ke2019predicting, Druz2020ManagersTone}. For example, \cite{tetlock2007giving} showed that media pessimism can predict downward pressure on stock prices and higher trading volume, while sentiment signals extracted from investor discussions and social media have also been linked to market direction \cite{das2007yahoo, bollen2011twitter}. \cite{loughran2011liability} showed that these signals can be improved using finance-specific dictionaries over general sentiment lexicons. 

In recent years, these ideas have been extended using transformer-based models, especially LLMs. \cite{araci2019finbert} adapted general-context BERT into FinBERT, specifically for financial sentiment analysis; \cite{wu2023bloomberggpt} proposed BloombergGPT, demonstrating the value of finance-specific large-scale pretraining of LLMs; and \cite{lopezlira2023chatgpt} showed that the LLM-based sentiment extracted from financial headlines contains predictive information for near-term market returns. Notably, much of this literature treated text as document-level or event-level signals, and lesser attention has been given to internal structure of expert discussions. 

\subsection{Expert Communication and Committee Deliberation}
Another stream of existing literature explores financial narratives. This includes earnings calls, analyst communications, and corporate reports, which usually contain qualitative information beyond numbers. \cite{Druz2020ManagersTone} studied influence of changes in tone during earnings calls over analysts and investors. Some studies showed that linguistic cues in conference calls can help detect deceptive communication or future financial risk \cite{larcker2012detecting, wang2014semiparametric}. These studies revealed that expert language can encode soft information, potentially relevant to financial outcomes. 

The closest work to this paper comes from the analysis done for central bank communications. \cite{hansen2018transparency} used computational linguistics to study deliberation within the Federal Open Market Committee (FOMC). \cite{osowska2024predicting} showed that the topical and sentiment features from the FOMC post-meeting statements help predict market reactions. These papers effectively demonstrated that committee language can contain meaningful signals worth exploring, however their focus has primarily been on public monetary-policy communications.

\subsection{Portfolio Management and Soft Information}
Portfolio-related NLP usually relies on external text sources such as news, social media, company descriptions, etc. Existing studies have examined how NLP-based forecasting and sentiment signals can support asset allocation decisions and portfolio construction \cite{xing2018natural, malandri2018public}. Some have used news-derived embeddings and semantic representation of firms to improve portfolio optimization \cite{xing2019growing, du2020stock}. The more recent work of \cite{hung2024intelligent} also includes news sentiments into portfolio construction processes. 

Some studies on soft-information emphasized the inclusion of qualitative and judgment based information that is difficult to encode in standard quantitative variables \cite{liberti2019information, wang2020screening}. The challenge of transforming deliberative information into structured features that are both predictive and interpretable still remains. 

\subsection{Structured and Context-Aware Representations}
Textual data can be represented in different ways: using lexicons, topic models, neural embeddings, and LLM-derived features, etc. Topic models such as Latent Dirichlet Allocation identify latent themes in documents \cite{blei2003latent}, while sentence-level embeddings provide dense semantic representation for classification and retrieval \cite{reimers2019sentencebert}. AutoML frameworks further provide the capability to combine tabular, textual, and embedding-based features efficiently \cite{erickson2020autogluon, tang2024autogluon}. These methods are powerful, but they often reduce interpretability since the information of which financial concepts drove the prediction gets lost. 

Taken together, the existing literature leaves several gaps relevant to our setting. Financial NLP has largely focused on public or corporate text and often treats text as a document- or event-level signal, while prior studies of committee deliberation have focused primarily on public monetary-policy communication. In addition, dense textual representations can obscure which financial concepts drive a prediction. CDSP addresses these gaps by using internal investment committee discussions as a signal source, conditioning financial language on the topical and asset-class context in which it occurs, and converting this structure into interpretable sentiment and attention features. We further compare these structured features with raw-text and embedding representations to assess whether interpretable and latent signals provide complementary predictive information.

\section{Data}
\label{sec:data}

\subsection{Meeting Transcripts}

In this paper, the data set is composed of transcripts of investment committee meetings usually convened on a monthly basis between June 2020 and February 2025. The committee actively discussed and deliberated macroeconomic conditions, specific asset-level outlooks for equities and bonds, and relative portfolio positioning (equities versus bonds). The meetings took place via video conference, hence transcripts created using a third-party software were made available. The dataset contained 48 transcripts due to uneven annual coverage. For example, year 2022 is over-represented (10 of 48 meetings) since the committee convened more frequently, likely due to post-COVID inflation.

\subsection{Target Variable}
In this paper, the prediction target is the \textit{binary market direction} for the month following each meeting. We define this binary market direction by whether or not global equities outperformed bonds in the subsequent calendar month:

\begin{equation}
y_t =
\left\{
\begin{array}{ll}
0  &\mbox{if }R^{\rm{eq}}_{t+1} \le R^{\rm{bond}}_{t+1} \\ 
1    &\mbox{if }R^{\rm{eq}}_{t+1} > R^{\rm{bond}}_{t+1},
\end{array}
\right.  
\end{equation}
where $R^{\rm{eq}}_{t+1}$ and $R^{\rm{bond}}_{t+1}$ are the next-month returns for equities and bonds. We used the Vanguard Total World Stock ETF (VT) and the Vanguard Total Bond Market ETF (BND) as global stock and bond proxies. 
In the 48-month dataset, equities outperformed in 29 months (60.4\%), establishing the majority-class baseline. The distribution reflects the broadly equity-positive period of 2020--2025, where stocks outperformed bonds every year, except in 2022.

\section{The CDSP Framework}
\label{sec:pipeline}

\begin{figure*}[t]
    \centering
    \includegraphics[width=\textwidth]{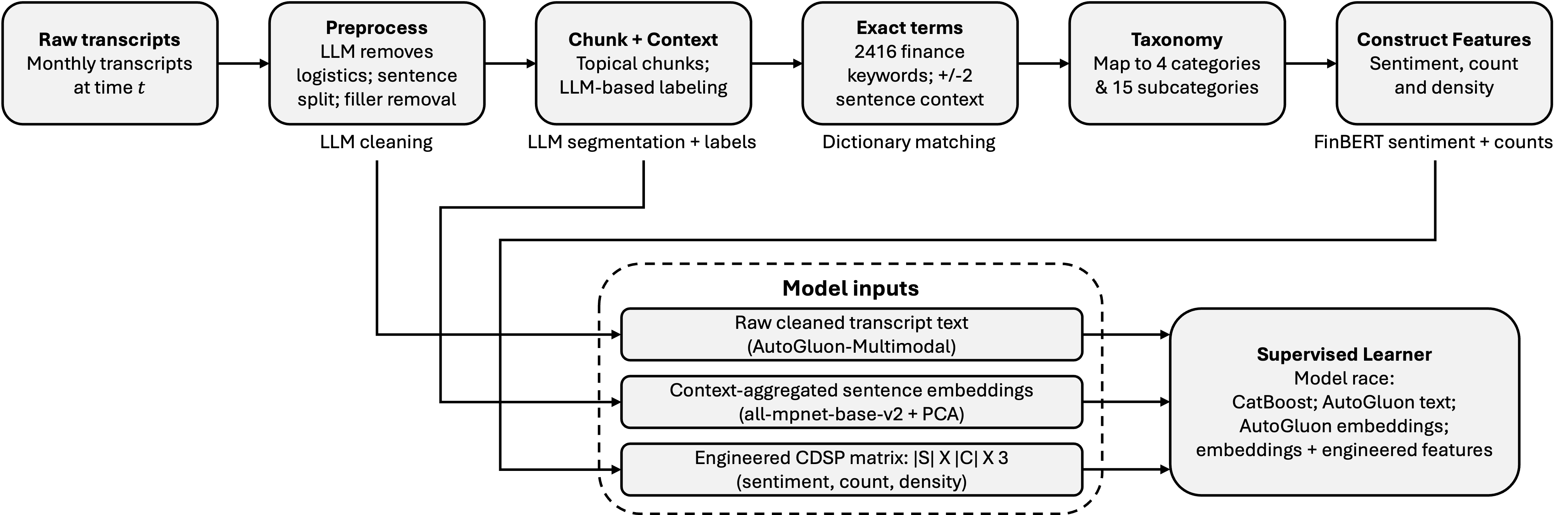}
    \caption{CDSP pipeline for converting investment committee transcripts into context-conditioned features and predictive model inputs}
    \label{fig:pipeline}
\end{figure*}

Figure~\ref{fig:pipeline} provides an overview of the full pipeline, which consists of four stages: transcript preprocessing, chunk segmentation and context labeling, keyword extraction and taxonomy mapping, and feature construction.

\subsection{Transcript Preprocessing}

Raw transcripts contain substantial meeting logistics: greetings, audio-check exchanges, scheduling discussion, and speaker transitions. These segments are identified and removed by an LLM classifier prompted to distinguish administrative from substantive content. The cleaned transcript is then split into sentences, and portfolio decisions (if any) and filler words (``um,'' ``you know,'' ``uh'') are removed. Portfolio decisions are then censored from the transcript to prevent the model from directly inferring the committee’s stated allocation decision. Apostrophes are preserved to maintain phrase integrity in the subsequent context window construction step. All LLM inferences hereafter are made using Claude Sonnet 4.5.

\subsection{Chunk Segmentation and Context Labeling}

The cleaned transcript is then segmented into topical chunks, each representing a coherent discussion segment identified by a start indicator, that is, the first substantive sentence of the segment, extracted by the same LLM used for preprocessing. Typical chunks correspond to single topics such as ``US equity valuation outlook'' or ``monetary policy trajectory.'' Each chunk is then assigned one or more context labels from the set: $\mathcal{C}=$\{equity, bonds, cross-asset, macroeconomic-outlook, meeting-logistic, other\}.

Labeling is performed by an LLM prompted with label definitions and few-shot examples. The multi-label design reflects the comparative nature of cross-asset deliberation: a chunk discussing equity valuations relative to bond yields receives both \textit{equity} and \textit{cross-asset} labels.

Subsequently, for context-conditioning: a keyword occurrence in an \textit{equity}-labeled chunk contributes to equity-context features, while the same keyword in a \textit{macroeconomic-outlook} chunk contributes to macro-context features. These are treated as entirely distinct features. Chunks labeled \textit{meeting-logistic} are excluded from all feature computation.

\subsection{Keyword Extraction and Taxonomy Mapping}

We maintain a vocabulary of 2,416 domain-specific financial keywords curated from investment committee language. The initial list of keywords is LLM-driven; the list is then cleaned up and consolidated with manual review. The vocabulary covers the terminology of cross-asset portfolio management: asset allocation concepts (``diversification,'' ``tactical positioning''), technical analysis terms (``yield curve,'' ``spread tightening''), economic outlook language (``soft landing,'' ``rate cycle,'' ``geopolitical risk''), and risk management terminology (``duration,'' ``hedging costs''). For each transcript, we record every occurrence of each vocabulary keyword, the chunk in which it appeared, and the chunk's context label(s). A context window of $\pm 2$ sentences around each keyword is then constructed for sentiment scoring.

Keywords are mapped to a 4-category, 15-subcategory taxonomy (given in Table~\ref{tab:taxonomy}), constructed iteratively through domain expert review. The taxonomy enables dimensionality reduction: rather than 2,416 individual keyword features, we aggregate to 15 subcategory-level signals.

\begin{table}[t]
\caption{Keyword Taxonomy: Categories and Subcategories}
\label{tab:taxonomy}
\centering
\begin{tabular}{ll}
\toprule
\textbf{Category} & \textbf{Subcategories} \\
\midrule
\multirow{2}{*}{Portfolio Management} & Asset Allocation \\
    & Strategic Framework \\
\midrule
\multirow{4}{*}{Technical Analysis}   
    & Corporate Analysis \\
    & Credit Analysis \\
    & Comparative Analysis \\
    & Quantitative Analysis \\
\midrule
\multirow{7}{*}{Economic Outlook}     & Economic Cycles \\
    & Economic Indicators \\
    & Forecasting \\
    & Macroeconomic \\
    & Monetary Policy \\
    & Market Dynamics \\
    & Geopolitical Factors \\
\midrule
\multirow{2}{*}{Risk Management}      & Macro Risk \\
    & Risk Measurement \\
\bottomrule
\end{tabular}
\end{table}

\subsection{Feature Construction}

For each transcript month $t$, subcategory $s \in \mathcal{S}$ (15 subcategories), and asset-class context $c \in \mathcal{C}$, three feature types are constructed:

\begin{itemize}
    \item \textbf{Sentiment score} $\sigma(t, s, c)$: the mean FinBERT\footnote{ProsusAI/finbert initialized through HuggingFace's AutoModel (batch size: 32, max length: 512, truncation: true, padding: true), \url{https://huggingface.co/ProsusAI/finbert}} \cite{araci2019finbert} sentiment of all keyword context windows in subcategory $s$ within context $c$ in month $t$, normalized from the raw positive/negative/neutral probability outputs to a scalar in $[0, 1]$ (higher = more positive sentiment).
    \item \textbf{Mention count} $\phi(t, s, c)$: the total keyword occurrences in subcategory $s$ within context $c$ in month $t$.
    \item \textbf{Mention density} $\delta(t, s, c)$: $\phi(t,s,c)$ normalised by the total word count of chunks assigned to context $c$ in month $t$, capturing relative discussion emphasis rather than absolute volume.
\end{itemize}

These features provide a month-level representation of each transcript that captures what topics were discussed, the contexts in which they appeared, and whether the surrounding language was positive or negative. For the classification experiments reported in this paper, we use only contemporaneous transcript-derived features from month $t$ to predict the market direction in month $t+1$.

\subsection{Feature Modalities}

Three modalities are used: (a) engineered features, (b) transcript text, and (c) sentence embeddings. The \textit{engineered features} include the $\{\sigma,\phi,\delta\}$ variables described in the previous subsection. The \textit{transcript text} includes full cleaned transcript text passed as a raw string to AutoGluon's\footnote{\url{https://auto.gluon.ai/dev/api/autogluon.multimodal.MultiModalPredictor.html}} multimodal predictor, which applies a transformer-based encoder for representational learning. The \textit{sentence embedding} features include 768-dimensional chunk-level embeddings (computed using a sentence transformer \cite{reimers2019sentencebert}, specifically \texttt{all-mpnet-base-v2}) aggregated to monthly vectors by context label. Principal component analysis (PCA) is then applied to reduce the overall dimensionality before training.

\section{Feature Description}
\label{sec:features}

\subsection{Mention Frequency vs. Sentiment as Predictors}

Table~\ref{tab:correlations} reports point-biserial correlations ($r_{pb}$) of mention count, mention density, and sentiment score features with the binary market direction target ($y_t = 1$: equity outperforms), for each subcategory. These are shown separately for ``All" (all chunks, regardless of context labels) and ``Equity" (restricted just to chunks with the equity context label). Bold values indicate the higher $|r_{pb}|$ between the two contexts. The large amount of text in these transcripts enables the CDSP approach. However, given the fact that the committee meets less than monthly, this limits the sample size ($N$ = 48) and the statistical significance of the correlations. 

\begin{table}[t]
\caption{Point-Biserial correlations ($r_{pb}$) with next-month market direction by subcategory and context ($N=48$). Bold: higher $|r|$ between all-context and equity-context. $^{*}p<0.05$.}
\label{tab:correlations}
\centering
\resizebox{\columnwidth}{!}{%
\begin{tabular}{lcccccc}
\toprule
 & \multicolumn{2}{c}{\textbf{Mention Count}} & \multicolumn{2}{c}{\textbf{Mention Density}} & \multicolumn{2}{c}{\textbf{Sentiment}} \\
\cmidrule(lr){2-3} \cmidrule(lr){4-5} \cmidrule(lr){6-7}
\textbf{Subcategory} & \textbf{All} & \textbf{Equity} & \textbf{All} & \textbf{Equity} & \textbf{All} & \textbf{Equity} \\
\midrule
Asset Allocation & \textbf{+0.020} & -0.018 & -0.067 & \textbf{-0.070} & +0.219 & \textbf{+0.294$^{*}$} \\
Strategic Framework & \textbf{+0.039} & -0.015 & \textbf{-0.038} & +0.000 & +0.070 & \textbf{+0.080} \\
\midrule
Corporate Analysis & -0.208 & \textbf{-0.218} & -0.088 & \textbf{-0.130} & -0.000 & \textbf{+0.037} \\
Credit Analysis & \textbf{+0.156} & +0.037 & -0.062 & \textbf{+0.092} & \textbf{+0.050} & -0.004 \\
Comparative Analysis & \textbf{+0.160} & +0.153 & \textbf{+0.180} & +0.131 & \textbf{+0.229} & +0.115 \\
Quantitative Analysis & +0.027 & \textbf{-0.037} & -0.118 & \textbf{-0.153} & +0.185 & \textbf{+0.191} \\
\midrule
Economic Cycles & +0.019 & \textbf{-0.141} & -0.013 & \textbf{-0.197} & \textbf{+0.384$^{*}$} & +0.264 \\
Econ. Indicators & +0.030 & \textbf{+0.047} & +0.003 & \textbf{-0.138} & \textbf{+0.174} & +0.164 \\
Forecasting & -0.079 & \textbf{-0.125} & \textbf{-0.367$^{*}$} & -0.180 & \textbf{+0.282$^{*}$} & +0.186 \\
Macroeconomic & +0.019 & \textbf{-0.228} & -0.072 & \textbf{-0.243} & \textbf{+0.241} & +0.131 \\
Monetary Policy & +0.005 & \textbf{+0.137} & \textbf{-0.059} & +0.009 & +0.049 & \textbf{+0.075} \\
Market Dynamics & \textbf{+0.101} & -0.086 & \textbf{-0.175} & -0.149 & \textbf{+0.224} & +0.062 \\
Geopolitical & -0.072 & \textbf{-0.284$^{*}$} & -0.165 & \textbf{-0.252} & \textbf{+0.158} & +0.021 \\
\midrule
Risk Measurement & +0.046 & \textbf{-0.067} & -0.090 & \textbf{-0.150} & \textbf{+0.160} & +0.082 \\
Macro Risk & \textbf{-0.156} & -0.108 & -0.116 & \textbf{-0.130} & +0.004 & \textbf{-0.023} \\
\bottomrule
\end{tabular}%
}
\end{table}

The sentiment feature displays stronger correlations with market direction than mention frequency (count and density)   across most subcategories. Among sentiment features, Economic Cycles, Asset Allocation, and Forecasting show the largest positive correlations, indicating that more positive deliberation sentiment in these categories precedes equity outperformance months. On the count channel, the strongest signals are negative: Geopolitical, Macroeconomic, and Corporate Analysis count features are associated with bond outperformance months.


\subsection{Context-Conditioning Effect}

Does context-conditioning improve the correlation with market direction? We see that this is the case more in the mention frequency features (count and density), but it seems to degrade correlation in the case of the sentiment feature. Therefore, context-conditioning appears to be helpful for some features, but not all. 


\subsection{Topic Attention Over Time}

\begin{figure}[t]
    \centering
    \includegraphics[width=\columnwidth]{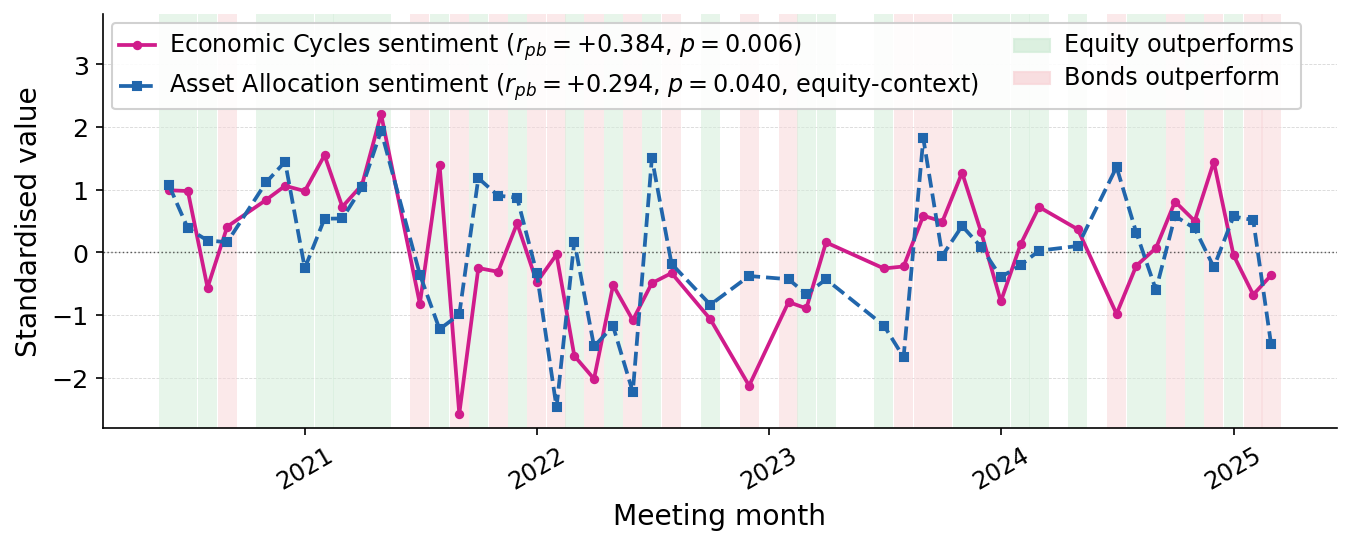}
    \caption{Standardised Economic Cycles sentiment ($r_{pb}=+0.384$, $p=0.006$) and Asset Allocation sentiment ($r_{pb}=+0.294$, $p=0.040$, equity-context) overlaid on next-month market direction outcomes (Jun 2020--Feb 2025). Green background: equity outperforms; red: bonds outperform.}
    \label{fig:topic_time_series}
\end{figure}

Figure~\ref{fig:topic_time_series} overlays the two strongest sentiment features, that is, Economic Cycles and Asset Allocation, against the monthly market direction outcome to illustrate how the top signals track the equity/bond regime over time. 

The two features confirm the pattern from Table~\ref{tab:correlations}: more positive deliberation sentiment in Economic Cycles and Asset Allocation subcategories tends to co-occur with subsequent equity outperformance. Neither feature tracks the outcome perfectly: both signals are noisy and the correlations are moderate. But the directional tendency is visible across the sample, including during the 2022 bond-outperformance episodes where both sentiment features are notably suppressed.

\section{Predictive Models}
\label{sec:model}

\subsection{Experimental Setup}

The model setup includes: (1) \textit{CatBoost} \cite{prokhorenkova2018} gradient boosted classifier with $L_2$ regularisation, trained on engineered semantic features only. (2) \textit{AutoGluon}\footnote{Model: AutoMM, \url{https://auto.gluon.ai/stable/tutorials/multimodal/multimodal_prediction/}} trained on raw transcript text only, which internally applies a transformer encoder to produce document representations. (3) AutoGluon\footnote{\label{autogluonensemble}Model: WeightedEnsemble, \url{https://auto.gluon.ai/stable/api/autogluon.tabular.models.html\#weightedensemblemodel}} trained on sentence embeddings alone. (4) AutoGluon\footref{autogluonensemble} trained on sentence embeddings combined with engineered features.

All models are evaluated under 5$\times$5 repeated stratified $k$-fold cross-validation. Each test has five repetitions with different random seeds to stabilise variance estimates on the small ($N=48$) dataset, producing a distribution of fold-level scores rather than a single estimate. 

Two baselines are used: (1) \textit{Majority-class}: always predict equity outperforms (60.4\% accuracy, 0.449 F1). (2) \textit{Lag-1 direction}: predict the same market outcome as the previous month (52.1\% accuracy, 0.494 F1). The lag-1 baseline underperforms majority-class accuracy because monthly outcomes have low serial correlation, that is, last month's winner is not a reliable predictor for this month.

CatBoost's hyperparameters are optimized using a Bayesian search paradigm\footnote{Utilizing Optuna to optimize learning rate: 0.03, depth: 6, L2: 6.9, and min data in leaf: 5, \url{https://optuna.readthedocs.io/en/stable/}}, while AutoGluon's hyperparameters are tuned using one of its preset modes\footnote{Presets: `best quality', \url{https://auto.gluon.ai/dev/tutorials/tabular/advanced/tabular-hpo.html}}.

\begin{table}[t]
\caption{Market direction classification results\\(5$\times$5 Repeated Stratified K-Fold, $N=48$)}
\label{tab:results}
\centering
\begin{tabular}{llcc}
\toprule
\textbf{Type} & \textbf{Model / Features} & \textbf{Acc.} & \textbf{F1} \\
\midrule
\multirow{2}{*}{Baseline}
    & Majority class         & 0.604 & 0.449 \\
    & Lag-1 direction        & 0.521 & 0.494 \\
\midrule
\multirow{4}{*}{Model}
    & CatBoost (Eng. features)        & 0.620 & 0.620 \\
    & AutoGluon (Transcripts)         & 0.710 & 0.690 \\
    & AutoGluon (Embeddings)          & 0.630 & 0.610 \\
    & AutoGluon (Embed.+Feat.)        & \textbf{0.730} & \textbf{0.730} \\
\bottomrule
\end{tabular}
\end{table}

\begin{figure}[t]
    \centering
    \includegraphics[width=\columnwidth]{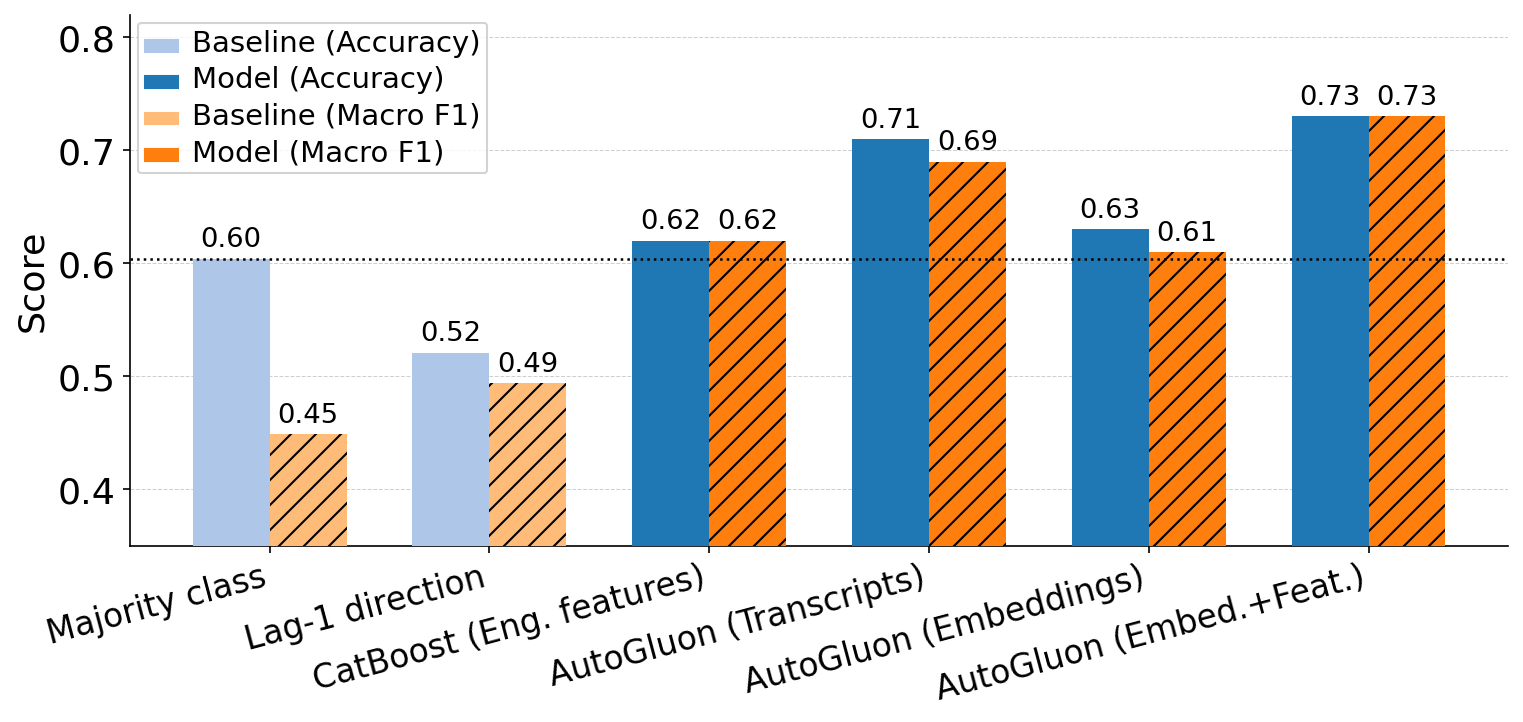}
    \caption{Accuracy (solid) and F1 (hatched) for all models and baselines. The dotted line marks the 60.4\% majority-class baseline. AutoGluon with combined embeddings and engineered features achieves a 12.6 percentage point improvement over the majority baseline.}
    \label{fig:model_results}
\end{figure}

\subsection{Results}

Table~\ref{tab:results} reports accuracy and F1 for all models and baselines, and Figure~\ref{fig:model_results} visualizes the comparison. The majority-class baseline strategy achieves 60.4\% accuracy but only 0.449 F1, reflecting near-zero recall on the minority (bonds-outperform) class. All four predictive models surpass this on F1, i.e., the transcript signal improves  minority-class identification over baseline. Furthermore, the lag-1 baseline is weak, achieving only 52.1\% accuracy, which is even below the majority-class baseline. This suggests that the predictive models, outperforming both baselines, are not just reflecting forward persistence in the target. Few other takeaways include: 
\begin{itemize}
    \item AutoGluon on transcripts alone achieves 71\% accuracy and 0.69 F1, without any feature engineering. This result suggests that the predictive signal is a property of the deliberation text, as opposed to an artifact of our specific taxonomy or feature construction choices.
    \item Combining sentence embeddings with engineered features lifts accuracy from 63\% to 73\% and F1 from 0.61 to 0.73. The engineered features capture structured, interpretable patterns (which subcategories are discussed, in which context, at what volume) that are not fully encoded in the dense embedding representation. The two modalities extract complementary information from the same source.
    \item CatBoost with engineered features only is the most interpretable model configuration. Without raw text or embeddings, the semantic features produce 62\% accuracy and 0.62 F1, thus improvement over the baselines is traceable to named subcategories.
\end{itemize}

\subsection{Feature Importance}

\begin{figure}[t]
    \centering
    \includegraphics[width=\columnwidth]{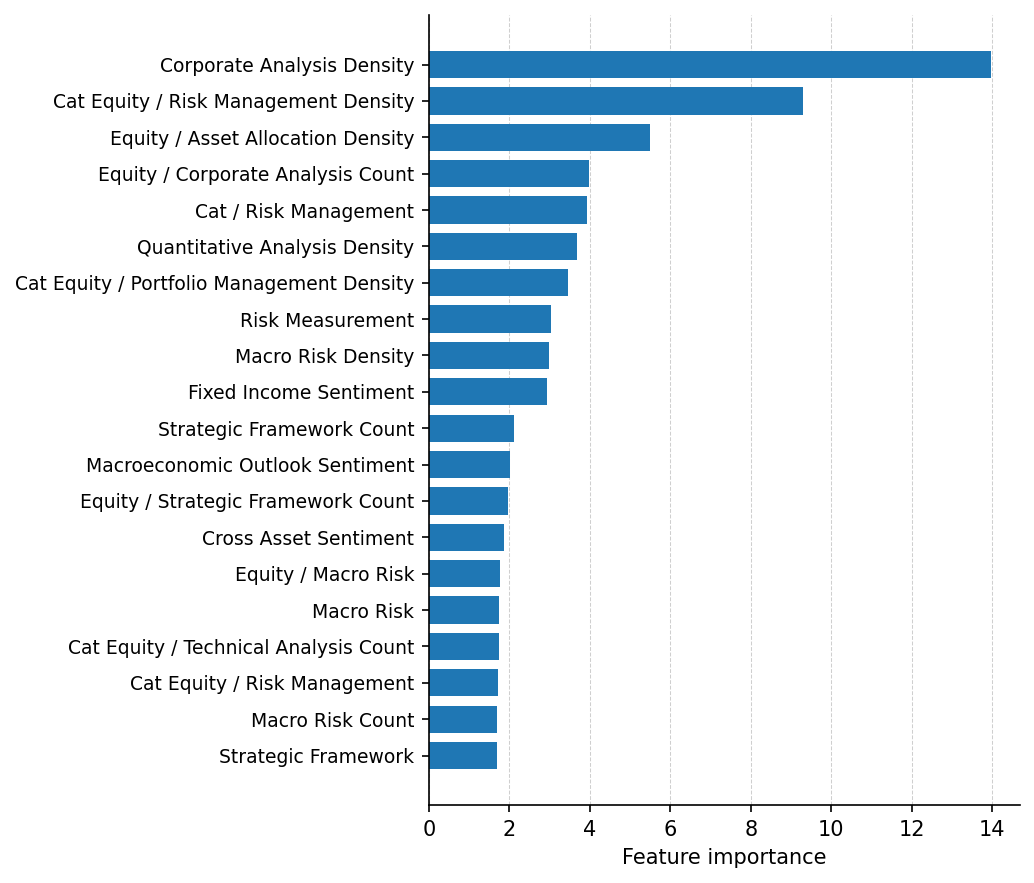}
    \caption{Top 20 CatBoost feature importances for market direction classification (engineered features only).}
    \label{fig:feature_importance}
\end{figure}

Figure~\ref{fig:feature_importance} shows feature importances from the CatBoost model trained on engineered features.
The most important feature is Corporate Analysis density, followed by equity-context Risk Management density and equity-context Asset Allocation density. The dominance of density features over raw count features is consistent with the correlation analysis: relative discussion emphasis is more informative than absolute keyword volume for some categories (see Table~\ref{tab:correlations}). The prominence of Risk Management and Asset Allocation features in equity-context specifically confirms that the same concept discussed during an equity-framed segment is more predictive than the same concept discussed in a general context, supporting the context-conditioning design of CDSP.

\section{Discussion}
\label{sec:discussion}

\subsection{Two Complementary Signal Channels}

The empirical results reveal two distinct and complementary signal mechanisms in the transcripts.

The first is \textit{structured deliberation signal}: what topics are discussed, how much attention they receive, and the sentiment with which they are discussed are all predictive of subsequent market direction. This channel is captured by the engineered features and is interpretable at the subcategory level. More positive Economic Cycles and Asset Allocation sentiment precedes equity outperformance; elevated Geopolitical and Macroeconomic discussion volume precedes bond outperformance. The sentiment channel dominates both count and density for several subcategories, while context-conditioning (equity-framed versus general) substantially strengthens the geopolitical and macroeconomic count/density signals.

The second is \textit{latent deliberation content}: the full transcript text, passed to a neural model without any feature engineering, achieves 71\% accuracy. This suggests the signal is richer than what the 15-subcategory taxonomy encodes: nuance in phrasing, speaker emphasis, the balance of hedged versus assertive language, and cross-sentence reasoning all contribute, even if they are not individually labeled. However, in regulated financial settings, predictive performance alone is insufficient: decisions must be explainable to risk managers, compliance officers, and investment committees. Engineered features satisfy this requirement directly: a prediction driven by elevated equity-context Geopolitical count or positive Economic Cycles sentiment can be traced back to specific discussion segments in the original transcript, whereas embedding-based models offer no easy attribution path.

The fact that combining both channels (73\%) improves on either alone (71\% and 63\%) suggests they are complementary rather than redundant. For practitioners, this suggests a two-layer architecture: an interpretable structured-feature model for explainability and a neural model for performance, combined at inference.

\subsection{Practical Implications}

The fact that the dual-channel method had a 73\% accuracy for our task of selecting the correct monthly equity-versus-bonds choice whereas the method of always selecting equities had only an accuracy of 60.4\% may be practically meaningful, but we also note this result is statistically weak. More specifically, applying McNemar's test to the same data yields a $p$-value of 0.263 (which, of course, is much higher than 0.05) largely due to the fact that we only have 48 data points. In a portfolio management context, better directional accuracy, applied to a binary risk-on/risk-off decision, may be useful despite market efficiency making prediction difficult, particularly in volatile regimes where the majority-class heuristic (i.e., always choosing equities) performs poorly. The finding that the signal persists across 2022 (a bond-outperformance year) and 2023--2024 (equity-outperformance years) suggests the model is capturing regime-varying features rather than a single-regime shortcut.

More broadly, the result supports the ``soft information'' hypothesis \cite{liberti2019information, wang2020screening}: qualitative deliberation encodes forward-looking information that is not already captured in the quantitative inputs available at the time of the meeting.

\subsection{Generalizability}

The CDSP pipeline is domain-agnostic. The four design elements (that is, chunk segmentation, LLM context labeling, taxonomy-structured extraction, and feature construction) transfer to any setting where experts deliberate over allocation or directional decisions in a transcribed, topically structured way. The only necessary contextual customization is creating the keyword taxonomy, which can be curated from the available dataset using any standard LLM model, and reviewed by domain experts for contextual relevance. While there are organic extensions of this framework to other domains in finance such as analyst calls deliberating stock related decisions, real-estate analysts deliberating construction pipelines, etc., there are possible extensions to non-finance domains as well, for example, management meetings. 

\subsection{Limitations}

\textbf{Dataset size:} $N=48$ observations is the primary constraint on all results. The 5$\times$5 repeated stratified CV mitigates fold-level variance, but each fold contains 38--40 training examples, with the risk of overfitting. The dominance of linear and simple tree models over deep ensembles in our experiments (AutoGluon's best models were typically linear) is consistent with this constraint.


\textbf{Transcript quality:} Automatic speech recognition introduces errors that reduce keyword recall, particularly for technical terms, acronyms (e.g., ``VIX,'' ``GDP,'' ``EPS''), and proper nouns. Estimated recall impact is 10--15\% of keywords in a representative set.

\textbf{Temporal structure:} The 5$\times$5 stratified CV does not preserve temporal ordering. A strict walk-forward evaluation, training on months 1--40 and evaluating on months 41--48, would be a more conservative out-of-time test, though with only 8 holdout points. Future work with a larger dataset should prioritize temporal evaluation.

\section{Conclusion}
\label{sec:conclusion}

This paper introduced CDSP, a context-conditional deliberation signal pipeline for transforming investment committee discussions into structured, predictive signals. The central idea is that expert discussion should not be treated as a flat text document. The same financial concept can carry different meaning depending on whether it appears in an equity discussion, a bond discussion, a macroeconomic overview, etc. CDSP preserves this structure by segmenting transcripts into topical chunks, assigning context labels, mapping keywords to a financial taxonomy, and constructing sentiment and attention-based features within each context. 

Applied to 48 investment committee transcripts, the framework produces two main findings. \textit{First}, sentiment polarity has higher correlations to next-month market direction than mention frequency. 
\textit{Second}, 
the transcript language contains some information beyond the engineered taxonomy. Raw transcript models perform well on their own, and the best result is achieved by combining embeddings with CDSP features, reaching an accuracy of 73\% against a 60.4\% majority-class (all equity) baseline over our 48 investment committee transcripts. 

The broader contribution is methodological. CDSP offers a way to convert qualitative expert reasoning into features that are both empirically testable and partially interpretable. This is important in financial settings where performance alone is insufficient and predictions must be traceable to recognizable topics, contexts, and discussion patterns. 

Overall, the evidence suggests that investment committee discussions may encode forward-looking information about market regimes. CDSP provides a practical way to recover that information, linking the qualitative structure of expert deliberation to supervised modeling. The same approach could be extended to other settings, such as buy-side/sell-side decisions, real-estate planning, medical discussions, and other expert forums where spoken reasoning is recorded and outcomes can be quantified.

\bibliographystyle{IEEEtran}
\bibliography{refs}

\end{document}